\documentclass[sigconf,screen]{acmart}

\usepackage{microtype}
\usepackage{graphicx}
\usepackage{subfigure}
\usepackage{xspace}
\usepackage{booktabs}   
\usepackage{hyperref}
\usepackage{multirow}   
\usepackage{soul}       
\usepackage{xcolor}
\usepackage{caption}
\usepackage{colortbl}   
\usepackage{pifont}
\usepackage{enumitem}
\usepackage{amsmath}
\usepackage{setspace}
\usepackage{csquotes}
\usepackage{adjustbox}
\usepackage[mathscr]{euscript}

\usepackage[capitalise]{cleveref}

\newcommand{\ilpc}{\textsl{\textsc{ILPC 2022}}\xspace}

\AtBeginDocument{%
  \providecommand\BibTeX{{%
    \normalfont B\kern-0.5em{\scshape i\kern-0.25em b}\kern-0.8em\TeX}}}

\setcopyright{rightsretained}
\copyrightyear{2022}
\acmYear{2022}
\acmDOI{XXXXXXX.XXXXXXX}

\acmConference[WWW22]{TheWebConf Workshop on Graph Learning Benchmarks 2022}{April 26, 2022}{Virtual}
\acmPrice{15.00}
\acmISBN{978-1-4503-XXXX-X/18/06}

\begin{document}

\title[An Open Challenge for Inductive Link Prediction on Knowledge Graphs]{An Open Challenge for Inductive Link Prediction\\on Knowledge Graphs}

\author{Mikhail Galkin}
\affiliation{%
  \institution{Mila \& McGill University}
  \city{Montreal}
  \state{QC}
  \country{Canada}
}
\email{mikhail.galkin@mila.quebec}
\orcid{0000-0003-3526-0155}

\author{Max Berrendorf}
\affiliation{%
  \institution{LMU Munich}
  \city{M\"unchen}
  \country{Germany}}
\email{berrendorf@dbs.ifi.lmu.de}
\orcid{0000-0001-9724-4009}

\author{Charles Tapley Hoyt}
\affiliation{%
  \institution{Harvard Medical School}
  \city{Boston}
  \state{MA}
  \country{USA}
}
\email{cthoyt@gmail.com}
\orcid{0000-0003-4423-4370}

\newcommand{\mycomment}[3]{\textcolor{#1}{[\bf #2: #3]}}
\newcommand{\cth}[1]{\mycomment{orange}{Charlie}{#1}}
\newcommand{\mg}[1]{\mycomment{green!70!black}{Michael}{#1}}
\newcommand{\mb}[1]{\mycomment{magenta}{Max}{#1}}

\renewcommand{\shortauthors}{Galkin, Berrendorf, and Hoyt}

\begin{abstract}
  An emerging trend in representation learning over knowledge graphs (KGs) moves beyond transductive link prediction tasks over a fixed set of known entities in favor of inductive tasks that imply training on one graph and performing inference over a new graph with unseen entities.
  In inductive setups, node features are often not available and training shallow entity embedding matrices is meaningless as they cannot be used at inference time with unseen entities. 
  Despite the growing interest, there are not enough benchmarks for evaluating inductive representation learning methods.
  In this work, we introduce \ilpc, a novel open challenge on KG inductive link prediction.
  
  To this end, we constructed two new datasets based on Wikidata with various sizes of training and inference graphs that are much larger than existing inductive benchmarks. We also provide two strong baselines leveraging recently proposed inductive methods.
  We hope this challenge helps to streamline community efforts in the inductive graph representation learning area.
  \ilpc follows best practices on evaluation fairness and reproducibility, and is available at \url{https://github.com/pykeen/ilpc2022}.
\end{abstract}




\maketitle

\section{Introduction}

The transductive link prediction (LP) task over knowledge graphs (KGs) has dominated the evaluation of knowledge graph embedding models (KGEMs) since their first appearance in 2011~\cite{DBLP:conf/icml/NickelTK11}.
Because the same graph is used for training and inference in the transductive setting, a majority of KGEMs use the \emph{shallow embedding} paradigm in which a learnable (or infrequently, a pre-computed) vector is assigned to each entity seen during training.
Most large benchmarking studies~\cite{ali2020benchmarking, Ruffinelli2020You, hu2020open} focus solely on the transductive LP task following the shallow embedding paradigm.

\begin{figure}[t]
    \centering
    \includegraphics[width=\columnwidth]{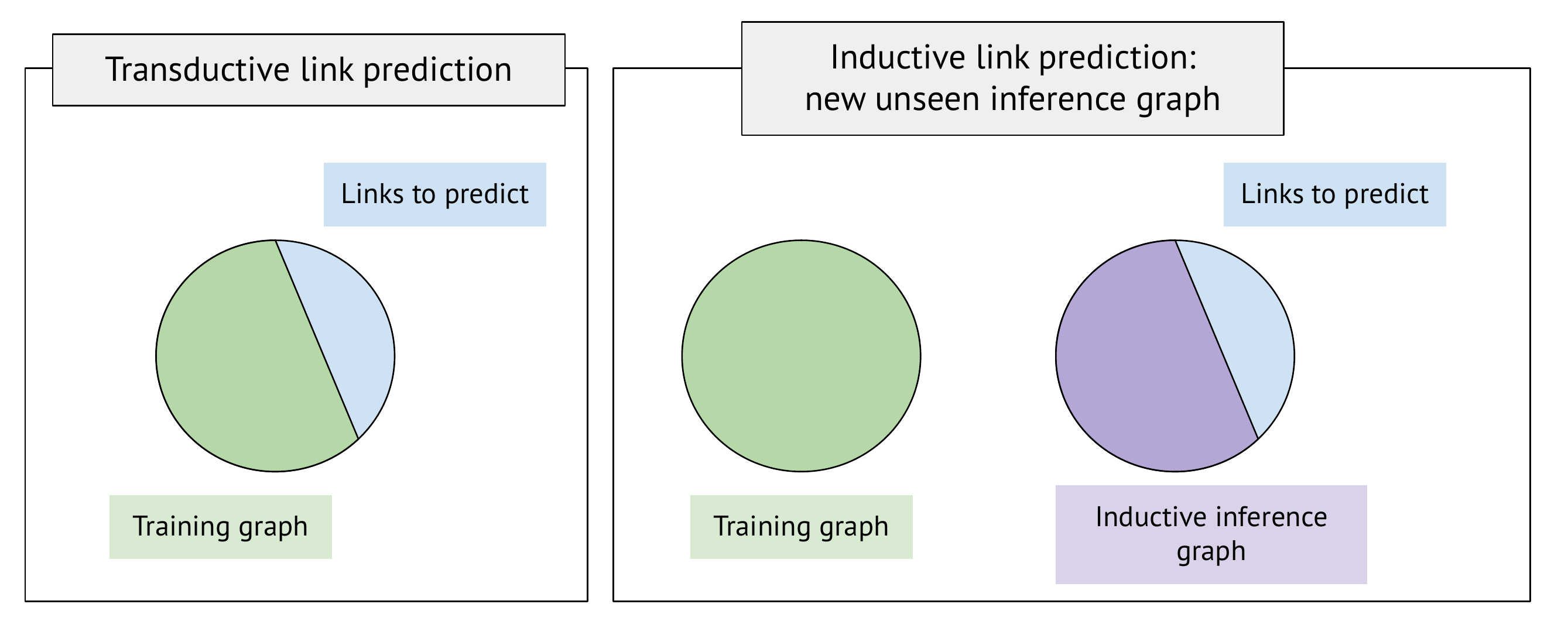}
    \caption{Transductive vs. inductive link prediction}
    \label{fig:inductive_lp1}
\end{figure}

In the inductive setting, a different graph can be used for inference that either contains a combination of seen and unseen entities (i.e., semi-inductive) or only unseen entities (i.e., fully inductive) and either the same or a subset of relations from the training graph~\cite{ali2021improving}.
During evaluation, there are two scoring tasks between entities seen and unseen during training: $\textit{seen} \leftrightarrow \textit{unseen}$ and $\textit{unseen} \leftrightarrow \textit{unseen}$.
In this work, we focus on the fully inductive setting where node features are not given and links to be predicted fall into the $\textit{unseen} \leftrightarrow \textit{unseen}$ category (\autoref{fig:inductive_lp1}).
Because the presence of unseen entities in the inference graph is fundamentally incompatible with the shallow embedding paradigm, several non-shallow methods compatible with the inductive setting have been recently proposed~\cite{teru2020inductive,zhu2021neural,galkin2022nodepiece}, many of which leverage recent advancements in graph neural networks and geometric deep learning~\cite{bronstein2021geometric}.

In this work, we outline the open Inductive Link Prediction Challenge (\ilpc) which aims to support community efforts to create more efficient and performant inductive link prediction models for KGs.
With it, we introduce two novel inductive link prediction datasets of different sizes leveraging Wikidata~\cite{vrandevcic2014wikidata} along with two strong baselines employing the recently proposed NodePiece~\cite{galkin2022nodepiece} representation learning method.

\section{Related Work}

There are far fewer high quality inductive link prediction benchmarks than transductive link prediction benchmarks.
For example, the Open Graph Benchmark~\cite{hu2020open} contains two datasets for transductive link prediction, \texttt{WikiKG2} and \texttt{WikiKG90Mv2}~\cite{hu2021ogblsc}, but none for inductive link prediction.
\texttt{Wikidata5M}~\cite{DBLP:journals/tacl/WangGZZLLT21} and inductive splits published in~\cite{daza2021inductive} demonstrate encoding entities by applying a language model to their respective descriptions. 
They evaluate inductive LP performance only on the single triple level without an inference graph. 
Therefore, those tasks can be seen as a probe of textual encoders and richness of their features rather than evaluation of graph representation learning capabilities. 

To the best of our knowledge, the only notable inductive LP benchmarks comprise inductive versions of three popular transductive LP datasets, \texttt{FB15k-237}~\cite{toutanova-chen-2015-observed}, \texttt{WN18RR}~\cite{DBLP:conf/aaai/DettmersMS018}, and \texttt{NELL-995}~\cite{chen2018variational} proposed in in~\cite{teru2020inductive}. 
Each KG accommodates four splits of various training and inference graph sizes making overall 12 distinct datasets.
Most of those datasets are rather small where the largest one has 27K training triples and 12K in the inference graph.
While the modest sizes were originally motivated by the computational complexity of the proposed method, some aspects have been resolved in subsequent works~\cite{zhu2021neural, galkin2022nodepiece}.

While our tasks draw inspiration from ~\cite{teru2020inductive}, the new \ilpc datasets are different in several ways: 1) the graphs are excerpts from Wikidata, the largest publicly available KG actively maintained by a large community with numerous practical applications~\cite{christmann2019look, decao2020autoregressive, decao2020multilingual}, hence, new results are more likely to be apprehended by a wider audience; 2) our largest graph is almost an order of magnitude larger than that of~\cite{teru2020inductive} and its inference graph size is challenging for modern GNN methods with inductive capabilities; 3) we make sure training and inference graphs are connected components without dangling nodes; 4) our evaluation criteria assumes ranking over all nodes in the inference graph instead of 50 random samples as done in~\cite{teru2020inductive}.
We describe the dataset construction process and design decisions in more detail in \cref{sec:ds_construction}.


\section{Task Definition}

Let
$
\mathbb{T}(\mathcal{E}, \mathcal{R}) := \{
(h, r, t) \mid h,t \in \mathcal{E}, r \in \mathcal{R}
\}
$
denote the set of possible triples between entities from $\mathcal{E}$ and relations from $\mathcal{R}$.
Let $\mathcal{E}_{\bullet}$ denote the entities occurring in training and $\mathcal{E}_{\circ}$ the entities in the inference graph.
Then, since we are in the fully inductive setting, we have a set of training triples $\mathcal{T}_{train} \subseteq \mathbb{T}(\mathcal{E}_{\bullet}, \mathcal{R})$ (comprising the \emph{training graph}), and a set of inference, validation and test triples 
$\mathcal{T}_{inf}, \mathcal{T}_{test}, \mathcal{T}_{validation} \subseteq \mathbb{T}(\mathcal{E}_{\circ}, \mathcal{R})$ where $\mathcal{T}_{inf}$ represents the \emph{inference graph}. 
Note that sets of training and inference entities are disjoint, i.e., $\mathcal{E}_{\bullet} \cap \mathcal{E}_{\circ} = \emptyset$.
The inductive link prediction task is defined as training a model on $\mathcal{T}_{train}$, running inference over a new graph $\mathcal{T}_{inf}$ and predicting missing links in the inference graph.

We use the two-sided workflow for evaluation that comprises both the head prediction and tail prediction task, i.e., for each evaluation triple $(h, r, t) \in \mathcal{T}_{eval}$, we consider the ranking tasks $(h, r, ?)$ and $(?, r, t)$, where we need to provide scores for each candidate entity $e \in \mathcal{E}_{\circ}$.
We use \textit{realistic} ranks, i.e., in case of ties, the assigned rank is the average of the best/smallest and the worst/largest rank in any ordering respecting the sort criterion~\cite{Berrendorf2020}.
As single-figure aggregations of the individual ranks, we use the following metrics, all sharing that larger numbers indicate better performance, and with an optimal value of 1.

\paragraph{\textbf{H@k}}
The Hits at $k$ (H@k) metric denotes the relative frequency of the correct entity's rank being at most $k$, for different $k\in\{1, 3, 5, 10, 100\}$.
Its value range is $[0, 1]$.

\paragraph{\textbf{MRR}}
The mean reciprocal rank (MRR) is the inverse of the harmonic mean of the individual ranks.
In contrast to H@k, which only considers the first $k$ entries, MRR considers all entries.
However, due to the reciprocal transformation, there are diminishing returns, and, inherited by its foundation on the harmonic mean, MRR mostly focuses on small ranks.
Its value range is $[0, 1]$.

\paragraph{\textbf{AMRI}}
The adjusted mean rank index (AMRI)~ \cite{Berrendorf2020} is an affine transformation of the (arithmetic) mean rank, which is normalized for the chance effect:
It is given as
$
AMRI = 1 - \frac{MR - 1}{\mathbb{E}[MR] - 1},
$
where MR denotes the arithmetic mean of ranks, and $\mathbb{E}[MR]$ its expectation under random ordering.
Thereby, it is implicitly normalized by the number of candidates, and thus allows comparison across cases with differing number of candidates, e.g., due to different number of entities, or filtered candidates.
Its value range is $[-1, 1]$, where 0 corresponds to a performance equal to the expected performance under random ordering of entities for the ranking task.

\section{Dataset Construction}
\label{sec:ds_construction}

The design principles behind creating \ilpc include that the dataset: i) should represent a real-world KG used in many practical downstream tasks; ii) should be challenging for modern inductive LP approaches and GNN architectures; iii) should be larger than existing benchmarks; iv) should allow for faster iteration and hypothesis testing; v) training and inference graphs should each be a connected component.

We chose CoDEx~\cite{safavi2020}, a high-quality subset of Wikidata for transductive link prediction, as a source of triples then sampled two inductive datasets then using the following procedure:
\begin{enumerate}
    \item Join the training, validation, and test triples of the original transductive dataset into a single graph.
    \item Sample a set of entities $\mathcal{E}_{\circ}$ for the inference graph. The remaining entities $\mathcal{E}_{\bullet}$ comprise the training graph.
    \item Sample subgraphs induced by $\mathcal{E}_{\bullet}$ (training graph) and $\mathcal{E}_{\circ}$ (inference graph) from the total set of triples $\mathcal{T}$.
    \item Fix the set of relations $\mathcal{R}$ in the training graph and ensure the relations in the inference graph are a subset of $\mathcal{R}$, otherwise remove the triple from the inference graph.
    \item Remove disconnected nodes from both training and inference graphs to ensure the graphs are connected components.
    \item Sample validation triples $\mathcal{T}_{validation}$ and test triples $\mathcal{T}_{test}$ from the inference graph $\mathcal{T}_{inf}$ such that they only contain nodes from $\mathcal{T}_{inf}$. 
\end{enumerate}

\begin{table}[]
    \centering
    \begin{tabular}{lrrr|rrr}
        \toprule
         & \multicolumn{3}{c}{ILPC22-S} & \multicolumn{3}{c}{ILPC22-L} \\
         \textbf{Split} & $|\mathcal{E}|$ & $|\mathcal{R}|$ & $|\mathcal{T}|$ & $|\mathcal{E}|$ & $|\mathcal{R}|$ & $|\mathcal{T}|$\\
         \midrule
         Training          & 10,230  & 96 & 78,616 & 46,626  & 130 & 202,446\\
         Inference         &  6,653  & 96 & 20,960 & 29,246  & 130 &  77,044\\
         Inference (val.)  &  6,653  & 96 &  2,908 & 29,246  & 130 &  10,179\\
         Inference (test.) &  6,653  & 96 &  2,902 & 29,246  & 130 &  10,184\\
         Inference (hold)  &  6,653  & 96 &  2,894 & 29,246  & 130 &  10,172\\
         \midrule
         Total & 16,883 & 96 & 108,280 & 75,872 & 130 & 310,045 \\
         \bottomrule
    \end{tabular}
    \caption{Statistics for ILPC22-S and ILPC22-L}
    \label{tab:dataset-statistics}
\end{table}

Following this procedure, we sample two datasets, ILPC22-S (small) and ILPC22-L (large). 
For both datasets, the size of the inference graph is about 66\% of the training graph in the number of entities and about 33\% of the training graph in the number of triples.
The sizes of the validation and test sets are about 10\% of the inference graph. 
While the validation and test sets are openly available for participants comprising a \emph{public} leaderboard, we also sample a \emph{hold-out test set} of triples from the inference graph $\mathcal{T}_{inf}$ for the final evaluation of model submissions. 
The hold-out test set is kept privately by the authors and is not publicly available in the challenge repository.

\begin{figure}[!h]
    \centering
    \includegraphics[width=\columnwidth]{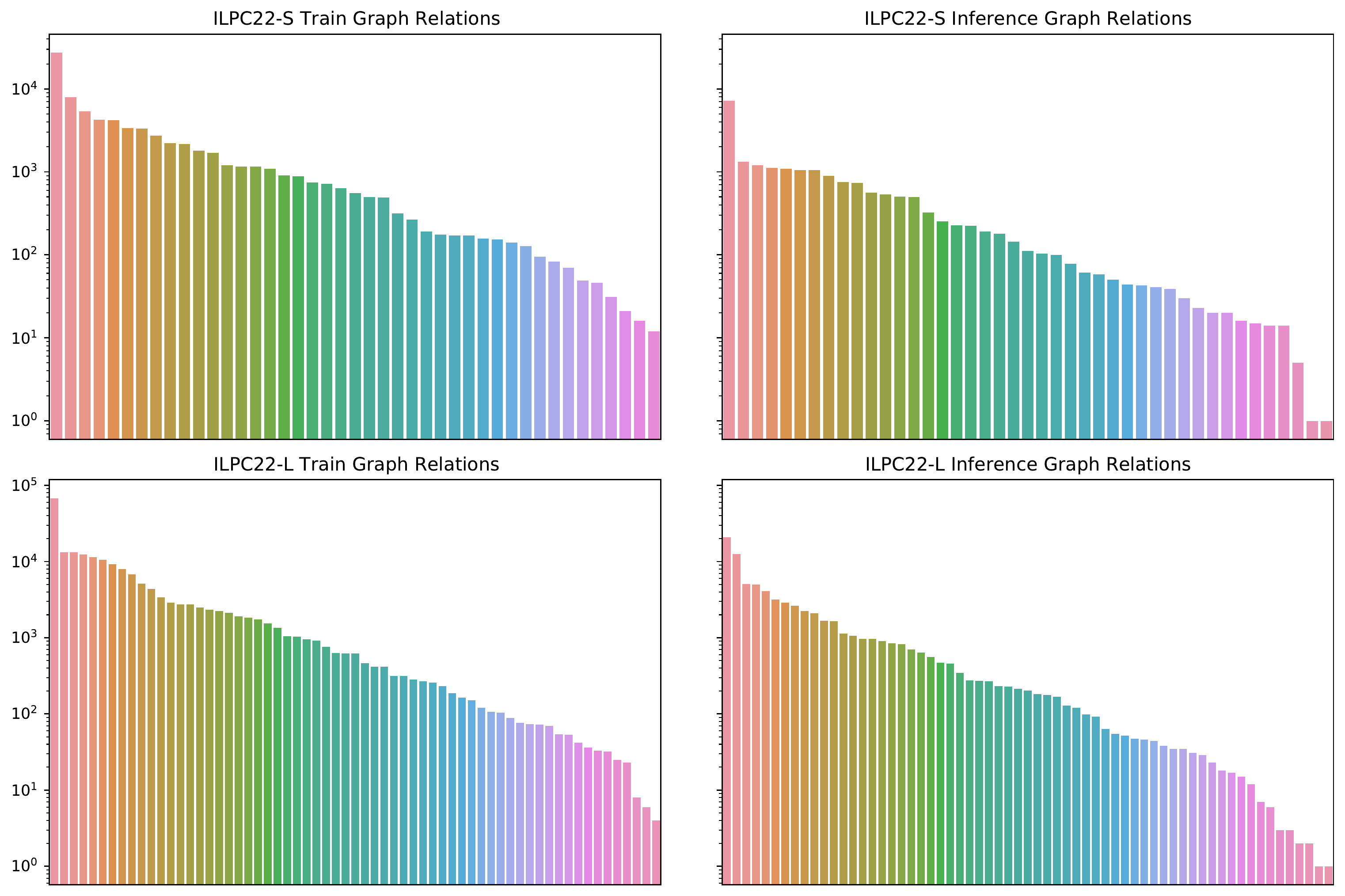}
    \caption{Distribution of relation type frequencies (without inverses) in training and inference graphs of ILPC22 datasets.}
    \label{fig:inductive_lp2}
\end{figure}

The smaller ILPC22-S is designed for faster hypothesis testing and model iteration and consists of 10K entities and 96 relations in the training graph with 6.6K entities in the inference graph.
The larger ILPC22-L is a challenging inductive dataset with 46K training and 30K inference entities with 130 relations, where the size of the inference graph (77K triples) is comparable to the whole training graph of smaller ILPC22-S. 
The overall statistics of ILPC22-S and ILPC22-L are presented in \autoref{tab:dataset-statistics} and \autoref{fig:inductive_lp2}.

We used the Zenodo integration with GitHub to assign a DOI to each release to maximize reproducibility. Users should report/cite the versioned DOI of the dataset they used for training along with their results, such as \cite{ilpc2022github}.

\section{Experiments and Baselines}
\label{sec:experiments}

We provide two strong baselines employing NodePiece \cite{galkin2022nodepiece}, a recently proposed inductive graph representation learning method that learns a fixed-size vocabulary of \emph{tokens} that remain invariant at training and inference time. 
This way, both seen and unseen nodes can be uniformly \emph{tokenized} using the same basic vocabulary and their representations are obtained via encoding the resulting \emph{tokens set}. 
With the fully inductive setup of \ilpc, the vocabulary consists of unique relation types, their inverses, and an auxiliary padding token making the vocabulary size $|\mathcal{V}| = 2 \cdot |\mathcal{R}| + 1$ constant to the number of entities or total triples in graphs. 
For all baselines, NodePiece is configured to tokenize each entity with 5 unique outgoing relations and use a 2-layer MLP as a token encoder.

The first baseline uses just the NodePiece tokenization with the MLP encoder to obtain entity representations to be sent to the scoring function. 
It is designed to be the faster method for hypothesis testing and model iteration achieving high recall at moderate computational costs.
The second baseline adds a message passing GNN encoder on top of the obtained NodePiece representations to further enrich entity and relation representations.
Here, we use a 2-layer CompGCN \cite{DBLP:conf/iclr/VashishthSNT20} relational GNN architecture with the DistMult \cite{DBLP:journals/corr/YangYHGD14a} composition function.
We denote the first plain NodePiece baseline as \texttt{-GNN} and the second GNN-enabled baseline as \texttt{+GNN}.

Both implemented baselines have the same internal dimension of 32 (we did not find any noticeable performance improvement when increasing the dimensionality) and use the same DistMult scoring  function (decoder).
All models were trained in the negative sampling setup with self-adversarial loss (NSSAL) \cite{DBLP:conf/iclr/SunDNT19} using the Adam optimizer~\cite{DBLP:journals/corr/KingmaB14} on a single RTX 8000 GPU. 
During training, models on ILPC22-S consumed at most 2 GB VRAM, and at most 3 GB on larger ILPC22-L.
Other hyperparameters are listed in \autoref{tab:hyperparams}.

\begin{table}[]
    \centering
    \begin{tabular}{lrrrr}
        \toprule
         & \multicolumn{2}{c}{ILPC22-S} & \multicolumn{2}{c}{ILPC22-L} \\
        \textbf{Parameter} & -GNN & +GNN               & -GNN & +GNN \\
        \midrule
        Dimension       & \multicolumn{4}{c}{32}\\
        Tokens per node & \multicolumn{4}{c}{5}\\
        GNN layers      & -         & 2	    & -     & 2\\
        Composition function & -    & DistMult & - & DistMult\\
        Scoring function & \multicolumn{4}{c}{DistMult}\\
        Loss function   & \multicolumn{4}{c}{Self-adversarial (NSSAL)}\\
        Margin          & 5.0       & 2.0	& 15.0  & 20.0\\
        Learning rate   & \multicolumn{4}{c}{0.0001}\\
        Epochs          & 50     & 50	& 17 & 53\\
        \# Negatives    & \multicolumn{4}{c}{16}\\
        \# Parameters   & 15.5K & 24K & 15.5K & 24K\\
        Training (minutes)   & 6     & 77	& 5 & 480\\
        \bottomrule 
    \end{tabular}
    \caption{Hyperparameters}
    \label{tab:hyperparams}
\end{table}

\begin{table}[]
    \centering
    \begin{tabular}{lrr|rr}
        \toprule
         & \multicolumn{2}{c}{ILPC22-S} & \multicolumn{2}{c}{ILPC22-L} \\
        \textbf{Metric} & -GNN & +GNN               & -GNN & +GNN \\
        \midrule
        MRR   & 0.0381 & 0.1326 & 0.0651 & 0.0705\\
        H@100 & 0.4678 & 0.4705 & 0.287  & 0.374\\
        H@10  & 0.0917 & 0.2509	& 0.1246 & 0.1458\\
        H@5   & 0.0500 & 0.1899	& 0.0809 & 0.0990\\
        H@3   & 0.0219 & 0.1396	& 0.0542 & 0.0730\\
        H@1   & 0.007  & 0.0763	& 0.0373 & 0.0319\\
        AMRI  & 0.666  & 0.730	& 0.646  & 0.682\\
        \bottomrule
    \end{tabular}
    \caption{Baseline evaluation of two variations of the NodePiece model on both datasets}
    \label{tab:baseline-results}
\end{table}

Experiments were conducted using PyKEEN~\cite{pykeen} and reported in  \autoref{tab:baseline-results}.
We observe that despite short training times, the plain NodePiece baselines (\texttt{-GNN}) already achieves relatively high recall in terms of H@100.
We attribute that to the effect of relation-based entity tokenization, i.e., a sample of incident relations is discriminative enough for broad separation of possibly correct links which is captured by H@100. 
While the GNN-enabled baseline exhibits 4-10$\times$ performance improvements ILPC22-S, it does not demonstrate such high gains for ILPC22-L, suggesting that the large dataset still poses challenges for state of the art GNNs and methodological improvements are still needed.

\subsection{Participation}

We solicit submissions from both individual researchers and teams, the submission instructions and leaderboards are provided in the official repository \url{https://github.com/pykeen/ilpc2022}. 

\section{Conclusion}

We presented \ilpc, a new open challenge on inductive link prediction on knowledge graphs. For this challenge, we sampled two new datasets from Wikidata larger than existing benchmarks. 
We make sure the target metrics are meaningful, interpretable, and follow the best reproducibility criteria.
For starters, we provided two baselines leveraging recently proposed inductive models that are fast to train on one hand or achieve higher performance on the other hand. 
We invite more submissions for this challenge extending the baselines or proposing new inductive LP models where the authors could estimate their performance on the public leaderboard. 
The challenge is expected to run for 6 months and we plan to report the intermediate results of submitted models on the private test set at the GLB workshop at the WebConf 2022.

\begin{acks}

Mikhail Galkin was supported by the Samsung AI grant held at Mila.
Max Berrendorf was supported by the German Federal Ministry of Education and Research (BMBF) under Grant No. 01IS18036A.
Charles Tapley Hoyt was supported by the DARPA Young Faculty Award W911NF20102551.
The authors of this work take full responsibilities for its content.

\end{acks}

\bibliographystyle{ACM-Reference-Format}
\bibliography{references}


\end{document}